\documentclass{article}
\usepackage{spconf,amsmath,amssymb,epsfig}
\usepackage{color}
\usepackage{acronym}
\usepackage{hyperref}
\usepackage{tabularray}
\usepackage{graphicx}
\usepackage{multirow}
\usepackage{comment}
\usepackage[normalem]{ulem}
\usepackage{mathtools}
\usepackage{microtype} 
\usepackage{stfloats} 
\usepackage{soul}

\DeclareMathOperator*{\argmax}{arg\,max}

\newcommand*\rot{\rotatebox{90}}
\definecolor{Alto}{rgb}{0.93,0.93,0.93}

\acrodef{LPIS}{Land Parcel Identification System}
\acrodef{MGRS}{Military Grid Reference System}
\acrodef{LULC}{land-use/land-cover}
\acrodef{DL}{Deep Learning}
\acrodef{DNN}{Deep Neural Network}
\acrodef{CNN}{Convolutional Neural Network}
\acrodef{RNN}{Recurrent Neural Network}
\acrodef{LSTM}{Long-Short-Term Memory}
\acrodef{TempCNN}{Temporal CNN}
\acrodef{GRU}{Gated Recurring Unit}
\acrodef{TOA}{Top-of-Atmosphere}
\acrodef{HMM}{Hidden Markov Model}
\acrodef{TPE}{Tree-Structured Parzen Estimator}
\acrodef{SITS}{satellite image time series}
\acrodef{S2}{Sentinel-2}
\acrodef{TS}{time series}
\acrodef{TE}{Transformer Encoder}
\acrodef{FT}{fine-tuning}

\title{Bayesian Modelling of Multi-Year Crop Type Classification Using Deep Neural Networks and Hidden Markov Models}
\name{Gianmarco Perantoni, Giulio Weikmann, Lorenzo Bruzzone
\thanks{Proc. IEEE Int. Geosci. Remote Sens. Symp., 2024, pp. 941–945, Preprint, Full version: \href{https://doi.org/10.1109/IGARSS53475.2024.10642432}{10.1109/IGARSS53475.2024.10642432}.}
\thanks{© 20XX IEEE.  Personal use of this material is permitted.  Permission from IEEE must be obtained for all other uses, in any current or future media, including reprinting/republishing this material for advertising or promotional purposes, creating new collective works, for resale or redistribution to servers or lists, or reuse of any copyrighted component of this work in other works.}}
\address{Department of Information Engineering and Computer Science, University of Trento, Italy}

\begin{document}
%\ninept
%
\maketitle
\begin{abstract}
The temporal consistency of yearly land-cover maps is of great importance to model the evolution and change of the land cover over the years. In this paper, we focus the attention on a novel approach to classification of yearly satellite image time series (SITS) that combines deep learning with Bayesian modelling, using Hidden Markov Models (HMMs) integrated with Transformer Encoder (TE) based DNNs. The proposed approach aims to capture both i) intricate temporal correlations in yearly SITS and ii) specific patterns in multiyear crop type sequences. It leverages the cascade classification of an HMM layer built on top of the TE, discerning consistent yearly crop-type sequences. Validation on a multiyear crop type classification dataset spanning 47 crop types and six years of Sentinel-2 acquisitions demonstrates the importance of modelling temporal consistency in the predicted labels. HMMs enhance the overall performance and F1 scores, emphasising the effectiveness of the proposed approach.
\end{abstract}
\section{Introduction}
\label{sec:intro}
In the last decade, \ac{DL} has become more and more relevant in remote sensing data analysis, allowing the research community to move towards more complex challenges on larger scales. However, with single acquisitions, it is not possible to model many characteristics of the land covers that are strictly related to their temporal signatures, such as in the case of cultivation or blooming vegetation. When multitemporal acquisitions are available, the use of \ac{DL} allows the modelling of the phenological traits of the land cover \cite{ghamisi2019multisource}. The \ac{DL} architectures considered for multitemporal remote sensing usually discard the spatial information and focus on the spectral and temporal information. They can be divided into three categories: i) \acp{RNN}, ii) 1D \acp{CNN} and iii) Transformer networks. Among the \acp{RNN}, the \ac{LSTM} network showed high accuracies \cite{russwurm2017temporal}.  In \cite{pelletier2019temporal}, the authors proposed a \ac{TempCNN} that applies 1D convolutions to the temporal domain, outperforming methods based on random forests and \acp{RNN} with \acp{GRU}. Transformer, which are based on the concept of Self-Attention, showed excellent performance in the natural language processing domain, and they are currently the most studied type of \acp{DNN} for \ac{SITS} analysis. Rußwurm \textit{et al.} \cite{russwurm2020self} compared Transformers, \ac{LSTM} networks and \acp{TempCNN} on a crop type classification task with \ac{S2} multispectral \ac{TS}. They showed that all these models perform similarly on pre-processed data (\textit{i.e.}, with atmospheric correction, temporal selection of cloud-free observations, and cloud masking), whereas \ac{LSTM} networks and Transformers perform better on raw \ac{TOA} data.
\par
A typical application of multitemporal acquisitions is the analysis of \ac{SITS} over the years to produce yearly \ac{LULC} maps. This application requires that multitemporal \ac{LULC} products outline a realistic temporal evolution of each pixel. For this reason, the temporal consistency of multitemporal land-cover products is particularly important. A Bayesian approach to the classification of two images acquired over the same location was proposed by Swain in his pioneering work \cite{swain1978bayesian}. He specifically modelled the temporal correlation between the land cover at different acquisition times to relate the class posterior probabilities estimated on single acquisitions. Then, he proposed a cascade implementation that, starting from a single time step, uses the temporal correlation modelled by known land-cover transition probabilities to account for temporal consistency with the second time step. Bruzzone \cite{bruzzone1999neural} extended this work to a compound classification system, proposing automatic techniques for the estimation of the transition probabilities and generalising the system to multisensor data and multiple classification systems (e.g., KNNs, SVMs and shallow neural networks) \cite{bruzzone2004detection}. Extensions to more than two observations usually rely on the \ac{HMM} formulation \cite{bouguila2022hidden}, which allows modelling the temporal correlation through the transition probabilities between adjacent time steps.
Nowadays, the classification of multiple dates with multitemporal data and \ac{DL}-based architectures is still a challenge, and the modelling of the temporal correlation in the \ac{TS} of the labels predicted by the \ac{DNN} is understudied  in remote sensing. Few works \cite{yu2010roles,seide2011feature,li2013hybrid,li2018method,buchan2023hmm} investigated the combination of \acp{HMM} with \acp{DNN} for acoustic data classification and ECG signal analysis. In remote sensing land cover classification, some papers \cite{sedona2022automatic} tried to overcome the complexity of the problem by training an end-to-end \ac{DL} model, such as an \ac{LSTM} network, with a multiyear training dataset, defined by the automatic extraction of reliable training samples from existing land cover products. However, the temporal correlation of the labels at different years is not modelled.
\par
In this paper, we focus on the combination of \acp{DNN} with strategies for modelling the temporal correlation in the \ac{TS} of predicted labels, drawing inspiration from a cascade classification approach \cite{swain1978bayesian}. The  strength of our approach lies in the fusion of Bayesian modelling, specifically \acp{HMM}, with \acp{DNN} based on the \ac{TE}. By combining the strengths of these methodologies, we aim to capture both the intricate temporal correlations in multiyear land-cover labels, and the complex spectral patterns within yearly \ac{SITS}. This strategy is validated on a multiyear crop type classification dataset spanning over six years of \ac{S2} acquisitions. The results show the importance of modelling the temporal consistency of the labels predicted by \acp{DNN}, as well as the effectiveness of their combination with \acp{HMM}.

\begin{figure}[t]
  \centering
  \includegraphics[width=\linewidth]{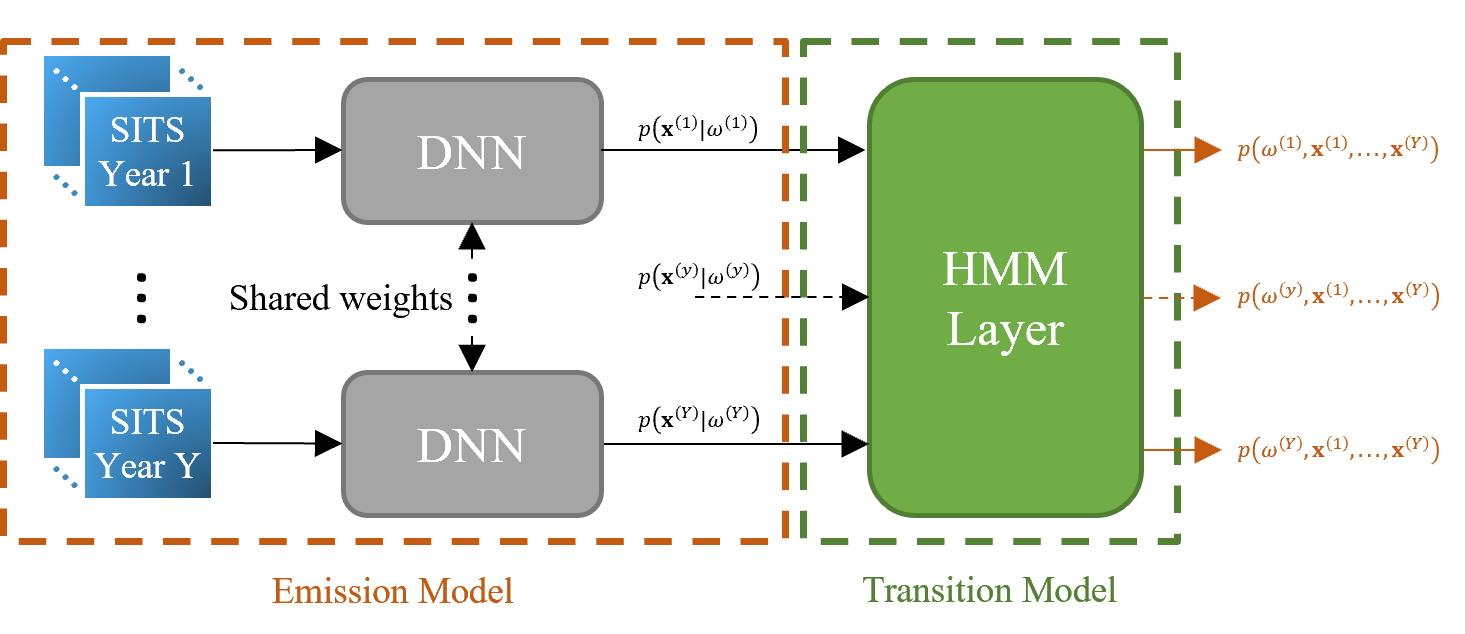}
  \caption{Flowchart of the proposed cascade-based classification approach, combining a DNN model with an HMM layer.}
  \label{fig:cascade}
\end{figure}

\section{Proposed Methodologies}
\label{sec:method}
In this section, we present the novel methodology based on cascade classification employed for multiyear crop type classification, drawing inspiration from \acp{HMM} and incorporating a \ac{TE} as the emission model (Fig. \ref{fig:cascade}). Let $\mathbf{x}^{\left(y\right)} \in \mathbb{R}^{T\times B}$ be a \ac{TS} of $T$ observations with $B$ spectral bands for year $y$. The task is to predict labels $\omega^{\left(y\right)} \in \left\{\omega_1,\omega_2,...,\omega_C \right\}, y=1,2,...,Y$, where $C$ is the number of classes and $Y$ is the number of years under consideration. Under a Bayesian perspective, the problem can be formulated as finding the best sequence of labels that maximise the joint density distribution:
\begin{equation} \small 
\begin{split}
    p\left(\omega^{\left(1\right)}, ..., \omega^{\left(Y\right)}, \mathbf{x}^{\left(1\right)}, ..., \mathbf{x}^{\left(Y\right)}  \right) = \\
    p\left(\mathbf{x}^{\left(1\right)}, ..., \mathbf{x}^{\left(Y\right)} \mid \omega^{\left(1\right)}, ..., \omega^{\left(Y\right)}  \right)P\left(\omega^{\left(1\right)}, ..., \omega^{\left(Y\right)}\right),
\end{split}
\end{equation}
where the two terms model the likelihood of the observed sequence given the generating sequence of labels and the prior probability of the sequence of labels, respectively.
In the cascade classification approach \cite{swain1978bayesian},  we consider class-conditional independence in the temporal domain,\textit{ i.e.}, the observations $\mathbf{x}^{\left(1\right)}, \mathbf{x}^{\left(2\right)}$ at a given spatial position at different time steps $y_1, y_2$ are independent of each other given the classes $\omega^{\left(1\right)}, \omega^{\left(2\right)}$ at the two time steps. Then, we can write as follows:
\begin{equation} \small 
\begin{split}
    p\left(\mathbf{x}^{\left(1\right)}, ..., \mathbf{x}^{\left(Y\right)} \mid \omega^{\left(1\right)}, ..., \omega^{\left(Y\right)}  \right) =\\
    \left[\prod_{y=1}^{Y}{p\left(\mathbf{x}^{\left(y\right)}\mid \omega^{\left(y\right)} \right)}\right],
\end{split}
\end{equation}
which is equivalent to the output independence assumption of \acp{HMM} \cite{bouguila2022hidden}. For two dates classification problems, the cascade classification approach \cite{swain1978bayesian} finds the best label $\omega^{\left(2\right)}$ by first marginalising $p\left(\omega^{\left(1\right)}, \omega^{\left(2\right)}, \mathbf{x}^{\left(1\right)}, \mathbf{x}^{\left(2\right)}\right)$ over $\omega^{\left(1\right)}$:
\begin{multline} \small
% \resizebox{\columnwidth}{!}{
    \argmax_i\biggl\{p\left(\mathbf{x}^{\left(2\right)}\mid \omega^{\left(2\right)}_i \right)\times \\
    \times\sum_{j=1}^C{P\left( \omega^{\left(2\right)}_i \mid \omega^{\left(1\right)}_j \right)p\left(\mathbf{x}^{\left(1\right)}\mid \omega^{\left(1\right)}_j \right)P\left(\omega^{\left(1\right)}_j \right)}\biggr\}.
% }
\end{multline}
One can note that this is akin to an \ac{HMM} where the Markov assumption is employed to simplify the definition and modelling of the prior term $P\left(\omega^{\left(1\right)}, ..., \omega^{\left(Y\right)}\right)$. Combining this with the \ac{HMM}'s output independence assumption and expanding to more than two dates, we perform cascade classification by using a custom implementation of an \ac{HMM}, where a \ac{DNN} is used to   approximately estimate the emission model $p\left( \mathbf{x} \mid \omega  \right)$ and a Markov transition model is used to enforce temporal consistency. 

\subsection{Emission Model: Transformer Encoder}
The proposed approach exploits a \ac{TE} for raw optical \ac{TS} classification \cite{russwurm2020self} as the emission model. This \ac{DNN} is designed to capture complex temporal and spectral patterns present in multitemporal \ac{SITS} data. Two versions of the \ac{TE} are considered, differing in the definition of the final layer: the first one uses a standard linear layer with softmax activation, while the second one uses a normalised softmax layer with no bias term and normalised rows in the weight matrix \cite{liu2017sphereface}. The former, considered as a discriminative approach,   approximately estimates class posterior probabilities $P_{disc}\left( \omega \mid \mathbf{x} \right)$ through standard categorical cross entropy optimisation, while the latter, akin to a generative model, implicitly enforces a uniform class prior distribution $P\left( \omega \right)=1/C, \forall \omega$ on the class posteriors, aligning with the requirements of an \ac{HMM} emission model, \textit{i.e.}, $ P_{gen}\left( \omega \mid \mathbf{x} \right) \sim p\left( \mathbf{x} \mid \omega  \right)$. This approach helps in reducing the bias towards dominant classes in the dataset. 

\subsection{Transition Model: HMM Layer}
To embody the \ac{HMM} logic, a custom layer is devised to serve as a unique implementation of the \ac{HMM} transition model for multiyear crop type classification. The parameters of this HMM layer include the log probabilities $P\left( \omega^{\left(y\right)}, \omega^{\left(y+1\right)} \right)$ of the joint prior distribution of each tuple of labels $\left(\omega^{\left(y\right)}, \omega^{\left(y+1\right)}\right)$, representing the probabilities of the two labels co-occurring in consecutive years. This configuration facilitates the extraction of both transition probabilities and initial state probabilities:
\begin{equation}  \small 
\begin{split}
    P\left( \omega^{\left(y+1\right)} \mid \omega^{\left(y\right)} \right) &= P\left( \omega^{\left(y+1\right)} , \omega^{\left(y\right)} \right)/P\left(\omega^{\left(y\right)} \right)\\
    P\left(\omega^{\left(y\right)} \right) &= \sum_{i=1}^C{P\left( \omega^{\left(y+1\right)}_i , \omega^{\left(y\right)} \right)},
\end{split}
\end{equation}
and improves numerical stability during both training and inference. Moreover, the modelling of the joint probabilities enables the estimation of both forward and backward transition probabilities, simply marginalising over $\omega^{\left(y\right)}$ instead of $\omega^{\left(y+1\right)}$. Leveraging this capability, the network performs cascade classification in both forward and backward directions during the inference step. This unique approach ensures that the predictions for each year depend on observations from all years, enhancing the overall consistency of predictions. The forward HMM model can be written as follows:
\begin{equation} \small
\begin{split}
    &p\left(\omega^{\left(1\right)}, ..., \omega^{\left(Y\right)}, \mathbf{x}^{\left(1\right)}, ..., \mathbf{x}^{\left(Y\right)}  \right) = \\
    &=\left[\prod_{y=1}^{Y}{p\left(\mathbf{x}^{\left(y\right)}\mid \omega^{\left(y\right)} \right)}\right]\left[\prod_{y=1}^{Y-1}{P\left( \omega^{\left(y+1\right)} \mid \omega^{\left(y\right)} \right)}\right]P\left(\omega^{\left(1\right)} \right).
\end{split}
\end{equation}
\par
\sloppy
In multiyear cascade classification,  the objective is to estimate for each time step $i$ the probability of classes $\omega^{\left(y\right)}$ occurring with the observed sequence up to that point:
\begin{equation} \label{eq:cascade} \small 
\begin{split}
    &p\left(\omega^{\left(y\right)}, \mathbf{x}^{\left(1\right)}, ..., \mathbf{x}^{\left(y\right)}  \right) =\\
    &= \sum_{i_j=1,\forall j < y}^C{p\left(\omega^{\left(1\right)}_{i_1},\omega^{\left(2\right)}_{i_2}, ..., \omega^{\left(y\right)}, \mathbf{x}^{\left(1\right)}, ..., \mathbf{x}^{\left(y\right)}  \right)}\\
    &
\begin{multlined}
    =p\left(\mathbf{x}^{\left(y\right)}\mid \omega^{\left(y\right)} \right) \sum_{i=1}P\left( \omega^{\left(y\right)} \mid \omega^{\left(y-1\right)}_i \right)\times\\\times p\left(\omega^{\left(y-1\right)}_i, \mathbf{x}^{\left(1\right)}, ..., \mathbf{x}^{\left(y-1\right)}  \right),
\end{multlined}
\end{split} 
\end{equation}
where the equation is applied recursively and $p\left(\mathbf{x}^{\left(1\right)}, \omega^{\left(1\right)} \right)=p\left(\mathbf{x}^{\left(1\right)}\mid \omega^{\left(1\right)} \right)P\left(\omega^{\left(1\right)} \right)$. Assuming that the observations for successive time steps are available, the same approach can be used in the backward direction, obtaining $p\left(\omega^{\left(y\right)}, \mathbf{x}^{\left(y\right)}, ..., \mathbf{x}^{\left(Y\right)}  \right)$. The forward and backward predictions can be fused, obtaining a prediction for $\omega^{\left(y\right)}$ that depends on the observation of all the years:
\begin{equation} \label{eq:fuse} \small 
\begin{split}
    &p\left(\omega^{\left(y\right)}, \mathbf{x}^{\left(1\right)}, ..., \mathbf{x}^{\left(Y\right)}  \right) =\\
    &=\frac{p\left(\omega^{\left(y\right)}, \mathbf{x}^{\left(1\right)}, ..., \mathbf{x}^{\left(y\right)}  \right)p\left(\omega^{\left(y\right)}, \mathbf{x}^{\left(y\right)}, ..., \mathbf{x}^{\left(Y\right)}  \right)}{p\left(\mathbf{x}^{\left(y\right)}\mid \omega^{\left(y\right)} \right)P\left(\omega^{\left(y\right)} \right)}.
\end{split} 
\end{equation}
This can be used both for training and inference, where class-posterior probabilities can be obtained simply by normalisation. In this paper, we also explore the second-order \ac{HMM}, where the transition probabilities are defined over combinations of the two previous time steps, \textit{i.e.}, we have $P\left( \omega^{\left(y+2\right)} \mid \omega^{\left(y+1\right)}, \omega^{\left(y\right)} \right)$. With some adjustments, the cascade equations can be formulated similarly to (\ref{eq:cascade}), and the results of forward and backward cascade predictions can be formulated exactly as in (\ref{eq:fuse}). Note that, differently from standard \acp{HMM}, we do not assume that the same transition model can be used for all the adjacent pairs of years, \textit{i.e.}, we estimate a transition matrix for each couple of years.
\subsection{Training Procedure}
The emission model is trained encompassing labels from all years. It is important to note that the \ac{TE} is specifically trained on individual yearly sequences rather than on the entire sequence collectively. The \ac{HMM} layer is initialised with co-occurrence matrices derived from the \ac{TE} training phase. The entire model, comprising the \ac{TE} and HMM layer, undergoes \ac{FT}, which is performed in a supervised manner by employing the cascade results of applying the forward-backward cascade algorithm in the categorical cross entropy.

\section{Dataset description}
\label{sec:dataset}
To validate the proposed methodologies, we considered publicly available \ac{LPIS} crop type maps in Austria, which are based on farmers' declarations\footnote{Data source: \url{https://www.data.gv.at/en/search/?typeFilter\%5B\%5D=dataset&searchterm=INVEKOS+Schl\%C3\%A4ge+\%C3\%96sterreich+&searchin=data}, Accessed on: May 23, 2024.}. The dataset has been developed with a focus on the \ac{MGRS} tile 33UVP. The selection of the tile was due to the consistent coverage of two \ac{S2} orbits and the wide variety of crop types present in the region. We analysed six target agronomic years, ranging from September 2016 to September 2022. The study involved a total number of 805 \ac{S2} L1C acquisitions, incorporating all the 13 spectral channels re-sampled at a resolution of $10m$ using a nearest neighbour approach. No cloud screening has been applied to the \ac{S2} images. For each agronomic year, we selected the crop fields with a minimum size of $400m^{2}$ and only the crop fields occurring in every target year with an occurrence greater than $0.05\%$.  At the end of this filtering process, we identified 47 distinct crop type labels, for a total number of ${\sim300,000}$ crop fields. Each resulting crop field is then collapsed into one single entry representing the spatial mean value of the pixels belonging to that crop field on every acquisition date.

\section{Experimental Setup and Results}
\label{sec:results}
To accurately model the temporal sequences and the consistency of the crop rotations, we split the dataset considering a stratified random sampling approach, considering the uniqueness of the label sequences within the agronomic target years. In particular, we defined each stratum exploiting an agglomerative hierarchical clustering approach, which recursively merges pairs of label sequences based on a Hamming distance with rotations. To determine the distance of two samples, we consider the minimum of each possible hamming distance computed on one sample and all the circular shifts of the other candidate sample. This approach results in the aggregation of samples showing similar crop sequences but shifted by one or more years, as they will display a small distance metric and be grouped within the same cluster. After the clustering operation, each sample in the stratum is randomly assigned to train, validation, and test sets, with a probability of $0.6$, $0.2$, and $0.2$, respectively. The \ac{TE} was trained on the training set, while the \ac{FT} of the \ac{TE} and \ac{HMM} was conducted considering the performance on the validation set. 

\begin{table}[t]
\centering
\caption{Comparison of the mean F1 scores (mF1\%) considering the different components of the methodology proposed on test set. Results from both training and \ac{FT} are reported.}
\label{tab:allcomparison}
\resizebox{0.8\linewidth}{!}{
\begin{tblr}{
  row{1} = {m,c},
  row{2} = {m,c},
  row{3} = {m,c},
  row{4} = {m,c},
  row{5} = {m,c},
  cell{1}{2} = {r},
  cell{2}{2} = {r},
  cell{3}{2} = {r},
  cell{4}{2} = {r},
  cell{5}{2} = {r},
  cell{6}{2} = {r},
  cell{7}{2} = {r},
  cell{2}{1} = {r=2}{},
  hline{1} = {1}{-}{},
  hline{1} = {2}{-}{},
  hline{2} = {1}{-}{},
  hline{2} = {2}{-}{},
  hline{4} = {-}{},
  hline{5} = {-}{},
  hline{6} = {1}{-}{},
  hline{6} = {2}{-}{},
}
                             & approach           & mF1\%        & FT mF1\% \\
TE                        & generative         & 68.71          & 70.88\\
TE                        & discriminative  & 64.85          & 70.05\\
{HMM \\1\textsuperscript{st} order}     &  cascade  & 70.16          & 73.59\\
{HMM \\2\textsuperscript{nd} order}   &  cascade  & 70.08          & 73.49\\
\end{tblr}
}
\end{table}
\begin{table*}[bp]
\centering
\caption{Quantitative analysis in terms of F1 scores (\%) on the 47 selected crop type classes comparing the fine-tuned generative \ac{TE} (GTE) and \acp{HMM}, considering both 1\textsuperscript{st} (HMM-1) and 2\textsuperscript{nd} (HMM-2) order cascade classification approaches. Best results are highlighted in gray.}
\label{tab:allclasses}
\resizebox{\textwidth}{!}{
\begin{tblr}{
  cells = {c},
  cell{2}{2} = {Alto},
  cell{4}{3} = {Alto},
  cell{4}{4} = {Alto},
  cell{4}{5} = {Alto},
  cell{4}{6} = {Alto},
  cell{3}{7} = {Alto},
  cell{3}{8} = {Alto},
  cell{3}{9} = {Alto},
  cell{2}{10} = {Alto},
  cell{2}{11} = {Alto},
  cell{4}{12} = {Alto},
  cell{4}{13} = {Alto},
  cell{3}{14} = {Alto},
  cell{4}{15} = {Alto},
  cell{3}{16} = {Alto},
  cell{4}{17} = {Alto},
  cell{4}{18} = {Alto},
  cell{4}{19} = {Alto},
  cell{4}{20} = {Alto},
  cell{3}{21} = {Alto},
  cell{4}{22} = {Alto},
  cell{4}{23} = {Alto},
  cell{3}{24} = {Alto},
  cell{4}{25} = {Alto},
  cell{4}{26} = {Alto},
  cell{3}{27} = {Alto},
  cell{3}{28} = {Alto},
  cell{3}{29} = {Alto},
  cell{4}{30} = {Alto},
  cell{4}{31} = {Alto},
  cell{4}{32} = {Alto},
  cell{3}{33} = {Alto},
  cell{4}{34} = {Alto},
  cell{4}{35} = {Alto},
  cell{4}{36} = {Alto},
  cell{2}{37} = {Alto},
  cell{3}{38} = {Alto},
  cell{3}{39} = {Alto},
  cell{4}{40} = {Alto},
  cell{3}{41} = {Alto},
  cell{3}{42} = {Alto},
  cell{2}{43} = {Alto},
  cell{3}{44} = {Alto},
  cell{3}{45} = {Alto},
  cell{3}{46} = {Alto},
  cell{3}{47} = {Alto},
  cell{3}{48} = {Alto},
  % vline{2} = {-}{},
  hline{1} = {1}{-}{},
  hline{1} = {2}{-}{},
  hline{2} = {-}{},
  hline{5} = {1}{-}{},
  hline{5} = {2}{-}{},
}
 \rot{Class Name} & \rot{WINTER BARLEY} & \rot{PASTURE THREE USES} & \rot{ PASTURE TWO USES } & \rot{ SILAGE MAIZE } & \rot{ WINTER TRITICALE } & \rot{ GREEN FALLOW } & \rot{ SINGLE-MOWN MEADOW } & \rot{ SUMMER OATS } & \rot{ BROAD BEANS } & \rot{ WINTER COMMON WHEAT } & \rot{ GRAIN MAIZE } & \rot{ WINTER MIXED CEREALS } & \rot{ TABLE POTATOES } & \rot{ DRESS } & \rot{ STRAWBERRIES } & \rot{ CLOVER } & \rot{ MISCELLANEOUS FIELD FODDER } & \rot{ FORAGE GRASSES } & \rot{ PERENNIAL PASTURE } & \rot{ ELEPHANT GRASS } & \rot{ GRASSLAND FALLOW } & \rot{ SPRING BARLEY } & \rot{ WINTER RAPESEED } & \rot{ SUMMER MIXED CEREALS } & \rot{ WINTER RYE } & \rot{ SUMMER COMMON WHEAT } & \rot{ SOYBEANS } & \rot{ WINTER SPELT } & \rot{ CORN COB MIX} & \rot{ ALTERNATING MEADOW } & \rot{ RANGELAND } & \rot{ OIL PUMPKIN } & \rot{ ENERGY WOOD } & \rot{ ALPINE FORAGE } & \rot{ TABLE APPLES } & \rot{ PEAS - CEREALS MIXTURE } & \rot{ OTHER ARABLE LAND } & \rot{ GRAIN PEAS } & \rot{ LUCERNE } & \rot{ WINTER CUMIN } & \rot{ SUMMER POPIES } & \rot{ SUNFLOWERS } & \rot{ SINGLE-CULTURE FIELD VEGETABLES } & \rot{ STARCH INDUSTRIAL POTATOES } & \rot{ SEED POTATOES } & \rot{ SUGAR BEETS } & \rot{ SEED MAIZE PROPAGATION}\\
GTE & 98.4 & 86.5 & 48.3 & 86.8 & 88.0 & 76.7 & 40.6 & 83.6 & 92.9 & 96.4 & 87.7 & 46.2 & 73.3 & 62.3 & 86.4 & 39.7 & 22.2 & 34.0 & 70.6 & 97.2 & 31.3 & 87.8 & 99.1 & 43.5 & 90.2 & 58.0 & 97.8 & 82.6 & 28.0 & 43.4 & 40.8 & 92.1 & 81.5 & 93.8 & 82.9 & 41.5 & 37.7 & 71.5 & 37.7 & 93.8 & 83.0 & 81.1 & 75.3 & 86.7 & 65.9 & 98.8 & 87.3\\
HMM-1  & 98.3 & 89.0 & 56.4 & 87.9 & 88.9 & 83.5 & 53.3 & 84.1 & 92.4 & 96.3 & 87.9 & 44.7 & 75.5 & 65.3 & 91.6 & 44.1 & 32.4 & 33.2 & 75.6 & 98.3 & 44.8 & 87.6 & 99.2 & 43.6 & 90.3 & 59.7 & 97.9 & 85.2 & 27.7 & 48.4 & 51.0 & 93.0 & 87.7 & 95.1 & 82.7 & 38.0 & 48.3 & 72.1 & 41.6 & 96.2 & 85.1 & 77.9 & 82.5 & 88.3 & 68.9 & 98.9 & 88.4\\
HMM-2 & 98.3 & 89.0 & 57.2 & 89.4 & 89.5 & 82.3 & 52.9 & 83.8 & 92.1 & 96.3 & 89.4 & 50.1 & 73.7 & 65.6 & 90.9 & 44.5 & 32.7 & 35.7 & 76.6 & 97.1 & 46.5 & 87.8 & 98.8 & 45.5 & 90.4 & 55.6 & 97.4 & 83.0 & 31.4 & 51.1 & 48.7 & 91.5 & 90.8 & 95.5 & 94.1 & 37.4 & 38.1 & 66.2 & 43.4 & 92.4 & 83.8 & 78.4 & 80.3 & 86.1 & 68.0 & 98.5 & 86.2
\end{tblr}
}
\end{table*}
\par
Table \ref{tab:allcomparison} shows a comparison of the performance of the different models on the test set, before and after \ac{FT}. \acp{HMM} before \ac{FT} are simply initialised with the co-occurrence matrices obtained from the training labels. From the table, one can see that the \ac{TE} alone performs worse compared to the combination with an \ac{HMM}, emphasising the importance of exploring the correlation in the temporal sequence. Moreover, the \ac{FT} cascaded \ac{HMM} achieves better performance compared to the inference-only approach (\textit{i.e.}, without \ac{FT}), showing a 73.59\% F1 score when considering the 1\textsuperscript{st} order approach. This is a notable achievement given the complexity of considering 47 different classes—which is a very challenging problem in crop-type mapping with remote sensing image classification. The modified version of the \ac{TE} employing the normalised weights in the final linear layer (\textit{i.e.}, generative) outperforms the original implementation of the \ac{TE} (\textit{i.e.}, discriminative) \cite{russwurm2020self}. A more detailed analysis can be inferred from Table \ref{tab:allclasses}, where the F1 scores are presented for each distinct crop type, considering the generative \ac{TE} and both $1^{st}$ and $2^{nd}$ order \acp{HMM}. Despite the similar performance on average, the classification results indicate that the $2^{nd}$ order \ac{HMM} performs better than the $1^{st}$ order \ac{HMM} on the most difficult classes at the expenses of the most performing ones.

\section{Conclusions}
\label{sec:conclusion}
In this paper, we have presented a novel cascade classification approach to multitemporal land cover classification of \ac{SITS} that integrates Bayesian modelling, specifically HMMs, with DNNs. This methodology involves two key aspects: (i) the supervised training of an \ac{TE} used as the emission model of an \ac{HMM} and (ii) the \ac{FT} of the cascade classification approach considering an \ac{HMM} model built on top of the Encoder. This study showcased that the incorporation of \ac{HMM} with \acp{DNN} effectively leverages the consistency of labels across different time sequences, outperforming the independent classification of each single year with the Encoder. The effectiveness of this novel strategy is validated through the experimental results on a multiyear dataset spanning six years of \ac{S2} acquisitions on a high detailed crop classification scheme, showing 47 different crop types. Our results highlight the importance of modelling the temporal consistency of labels predicted by \acp{DNN}, by fusing it with \acp{HMM} in a Bayesian approach, in enhancing the overall performance of crop type classification models.
In future works, we plan to reframe the approach within a change detection framework and investigate the weak and semi supervision scenarios, where only a subset of the training data are reliably labelled.

\bibliographystyle{IEEEbib}
\bibliography{strings,refs}

\end{document}